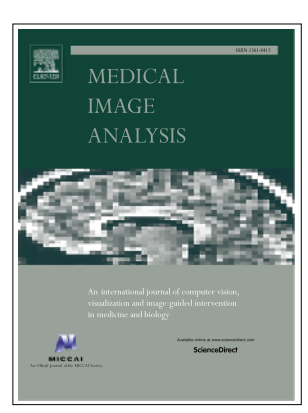

# Reason like a radiologist: Chain-of-thought and reinforcement learning for verifiable report generation

Peiyuan Jing [a,1], Kinhei Lee [a,1], Zhenxuan Zhang [a,1], Huichi Zhou [a], Zhengqing Yuan [b], Zhifan Gao [c], Lei Zhu [d], Giorgos Papanastasiou [e,f], Yingying Fang [a,g], Guang Yang [a,g,h,i,*]

[a] Bioengineering Department and Imperial-X, Imperial College London, W12 7SL, London, UK
[b] Department of Computer Science and Engineering, University of Notre, Holy Cross Dr, IN46556, Dame Notre Dame, USA
[c] School of Biomedical Engineering, Shenzhen Campus of Sun Yat-sen University, Guangzhou, China
[d] Robotics & Autonomous Systems Thrust & Data Science & Analytics Thrust, HKUST (GZ), Guangzhou, China
[e] Mathematics Research Centre, Academy of Athens, Greece
[f] Archimedes Unit, Athena Research Centre, Greece
[g] National Heart and Lung Institute, Imperial College London, SW7 2AZ, London, UK
[h] Cardiovascular Research Centre, Royal Brompton Hospital, SW3 6NP, London, UK
[i] School of Biomedical Engineering & Imaging Sciences, King's, WC2R 2LS, College London London, UK

ABSTRACT

Radiology report generation is critical for efficiency, but current models often lack the structured reasoning of experts and the ability to explicitly ground findings in anatomical evidence, which limits clinical trust and explainability. This paper introduces BoxMed-RL, a unified training framework to generate spatially verifiable and explainable chest X-ray reports. BoxMed-RL advances chest X-ray report generation through two integrated phases: (1) Pretraining Phase. BoxMed-RL learns radiologist-like reasoning through medical concept learning and enforces spatial grounding with reinforcement learning. (2) Downstream Adapter Phase. Pretrained weights are frozen while a lightweight adapter ensures fluency and clinical credibility. Experiments on two widely used public benchmarks (MIMIC-CXR and IU X-Ray) demonstrate that BoxMed-RL achieves an average 7 % improvement in both METEOR and ROUGE-L metrics compared to state-of-the-art methods. An average 5 % improvement in large language model-based metrics further underscores BoxMed-RL's robustness in generating high-quality reports. Related code and training templates are publicly available at https://github.com/ayanglab/BoxMed-RL.

## 1. Introduction

Radiological imaging plays a crucial role in clinical decision-making by significantly aiding disease diagnosis and patient management. The explainability and transparency of medical reports generated from radiological images are essential (Brady et al., 2012; Goergen et al., 2013). Inadequate explainability can lead to miscommunications among clinicians, potential misdiagnoses, decreased patient trust, and ultimately compromised patient safety (Satia et al., 2015). Clinicians rely on visual explanations to understand not only the diagnostic outcome but also the anatomical and pathological reasoning behind these conclusions (Pan et al., 2025). This requires models to align visual perception in images, such as identifying lesions, anatomical landmarks, and spatial relationships with clinical language (Rajpurkar and Lungren, 2023). The rapidly increasing volume of radiological images produced annually has created substantial workload pressures on radiologists, complicating thorough image interpretation and increasing the risk of diagnostic inaccuracies (Delrue et al., 2011). This issue has led to significant backlogs in radiological reporting, particularly within public healthcare systems, further intensifying the risk of medical errors and inefficient use of medical resources (Wang et al., 2022a; Yang et al., 2023; Gao et al., 2024). Therefore, there is a need for an explainable automated medical report generation model that combines fine-grained visual perception with clinical reasoning to generate reports (Fig. 1 (a)). Such tools require the ability to precisely localize and interpret visual findings (e.g., tumors, fractures, or tissue anomalies) while explicitly linking these observations to diagnostic conclusions in natural language. By doing so, they can efficiently summarize clinical findings while providing clinicians with transparent, image-grounded explanations that simulate the radiologist's analytical workflow (Tang et al., 2025).

* Corresponding author.
*E-mail address:* g.yang@imperial.ac.uk (G. Yang).
[1] These authors contributed equally.

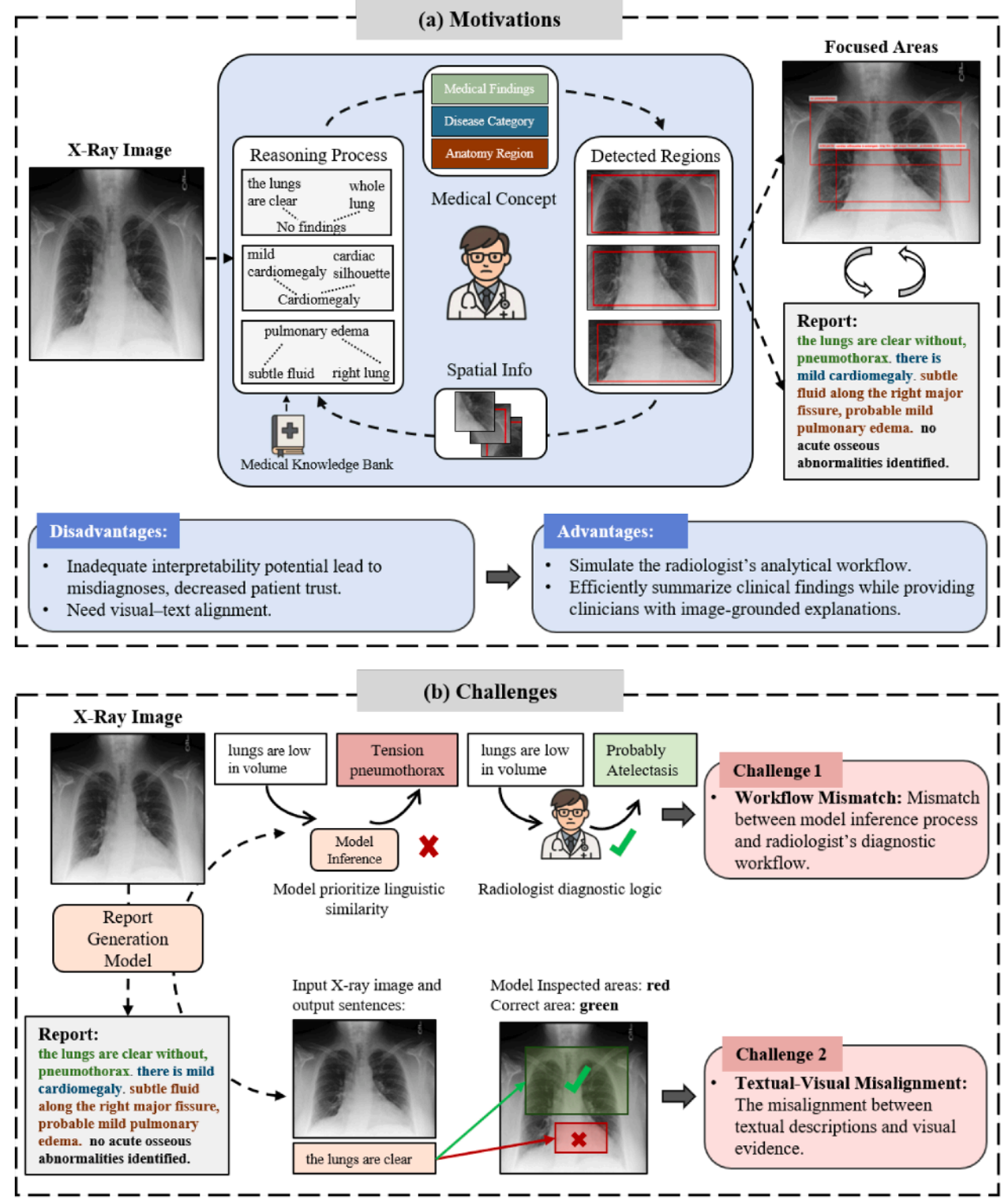

**Fig. 1.** Overview of motivations and challenges. (a) Clinical motivation: radiologists combine medical concepts (findings, disease categories, anatomical regions) with spatial clues when interpreting chest X-rays. (b) Existing challenges: workflow mismatch with radiologists' diagnostic logic and textual–visual misalignment between generated reports and image evidence.

Despite progress in automated medical report generation, existing methods face critical limitations in producing clinically trustworthy explanations (Wang et al., 2023a; Tanida et al., 2023). As shown in Fig. 1(b), automated report generation still faces two core challenges: (1) mismatch between model inference processes and radiologists' diagnostic workflows, and (2) misalignment between textual descriptions and visual evidence. First, most methods generate free-form narratives that deviate from radiologists' structured diagnostic logic (e.g., systematically analyzing anatomical regions to rule out differential diagnoses) (Rajpurkar and Lungren, 2023). These models lack the hierarchical reasoning ability to think like a radiologist; for example, they may state contradictory findings such as "lungs are low in volume" followed by "tension pneumothorax" (which typically causes increased lung volume) (Pan et al., 2025). Such errors arise because models prioritize linguistic similarity over clinical plausibility, failing to mimic the structured reasoning workflow of radiologists. Second, existing systems often often show limited ability to consistently correlate their statements with visual evidence (e.g., noting a "lung nodule" without localizing it) (Yang et al., 2022; Wang et al., 2022b). This creates ambiguity, as clinicians cannot verify whether findings (e.g., "normal pulmonary vasculature") are derived from image patterns or inferred from biased language priors (Zhang et al., 2024c; Tanida et al., 2023). For instance, a model might correctly describe "a lung nodule in the right upper lobe" but point to the left lung during inference. This occurs because current methods rely on tokenized visual representations that only capture implicit vision–language correlations rather than explicit anatomical understanding. As a result, models are allowed to "guess" plausible findings without spatial fidelity, eroding clinician trust and risking misdiagnosis (Satia et al., 2015; Zhang et al., 2024d).

Although recent methods have advanced report generation, they still struggle to resolve workflow mismatch and textual–visual misalignment

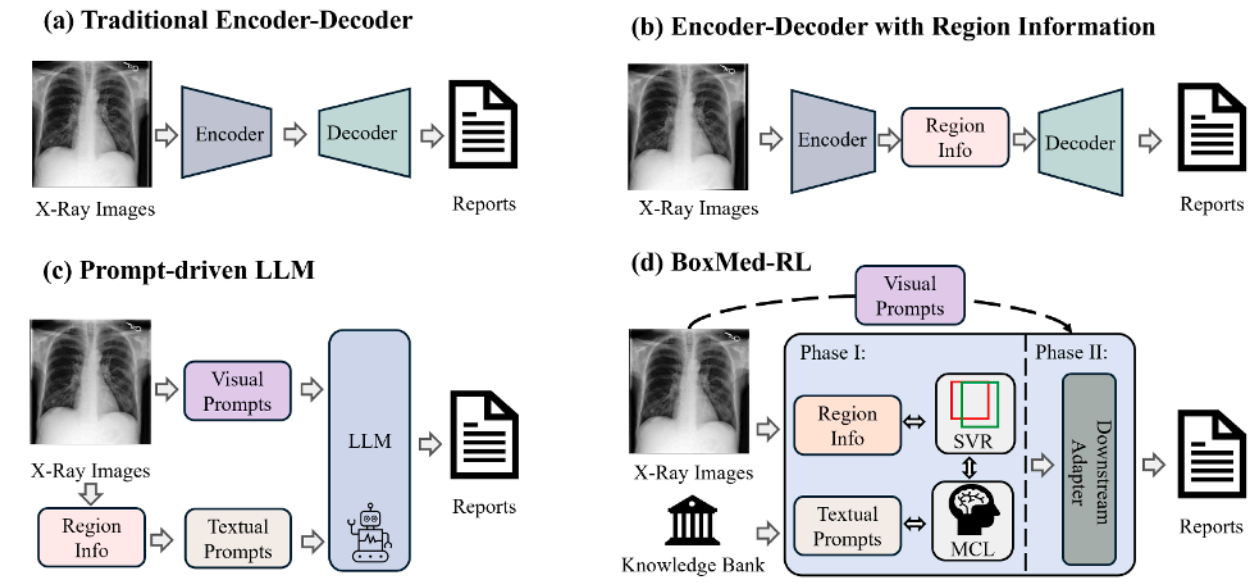

**Fig. 2.** Comparison of medical report generation architectures. (a) Traditional encoder-decoder. (b) Encoder–decoder with region information. (c) Prompt-driven large language models. (d) Our proposed BoxMed-RL.

(Chen et al., 2020; Liu et al., 2021a). **Traditional encoder–decoder models** (Fig. 2(a)), such as R2Gen (Chen et al., 2020) and GSK (Yang et al., 2022), generate free-form reports directly from input images without explicitly modeling spatial information. While effective for general captioning, this approach neglects the correspondence between findings and anatomical regions, limiting sensitivity to fine-grained image details. As a result, it is difficult to trace the model's reported findings back to the specific regions of the image from which they were derived. **Region–enhanced encoder–decoder models** (Fig. 2(b)), such as MA (Wang et al., 2022a) and CMN (Chen et al., 2021), mitigate this limitation by integrating regional features (e.g., lesion patches or organ-level segments) into the encoder. This enhances sensitivity to localized patterns and facilitates the detection of subtle or small-scale abnormalities. However, the overall reasoning process remains opaque: while the model can effectively associate visual patches with disease presence, it cannot explain why the disease manifests in those specific locations or why particular regions are diagnostically critical. **Prompt-driven large language models** (Fig. 2(c)), such as STREAM (Yang et al., 2025b) and R2GenGPT (Wang et al., 2023b), use textual and visual prompts to incorporate regional cues. However, spatial information is encoded in a limited way, leading to a loose alignment between vision and language. Although these models show moderate general reasoning by building upon large language models, they fail to replicate the structured diagnostic workflow of radiologists and often generate free-form narratives that lack clinical coherence. Across these categories, two critical flaws remain: (i) current architectures lack hierarchical diagnostic logic to validate their generated reports (e.g., from findings → disease → anatomy) and (ii) vision and language are treated as loosely coupled modalities, preventing spatially explicit grounding (e.g., linking "left lung nodule" to its bounding box). These flaws yield ambiguous or misaligned reports, which risk misdiagnosis and erode clinician trust (Sloan et al., 2024; Zhang et al., 2024d; Satia et al., 2015; Nan et al., 2025).

To address these critical challenges, we propose BoxMed-RL (Fig. 2 (d)), a two-phase framework aimed at improving explainability and clinical interpretability in automated medical report generation. Unlike prior methods that rely on free-form narratives or implicit vision–language correlations, BoxMed-RL explicitly integrates structured medical reasoning with spatially verifiable learning, enabling the model to generate reports that are both clinically coherent and anatomically grounded. BoxMed-RL advances explainable report generation through two key innovations in its Pretraining Phase: (1) Medical Concept Learning (MCL), a module to internalize the diagnostic flow of radiologist. (2) Spatially Verifiable Reinforcement (SVR), a module to ground textual descriptions in anatomical evidence. By applying the two innovations, BoxMed-RL incorporates domain-specific clinical concepts and textual-visual alignment ability, approximating the cognitive workflow of radiologists. The overall workflow of BoxMed-RL comprises the following: In the Pretraining Phase, our model is first fine-tuned via MCL using a structured radiological Chain-of-Thought (CoT) reasoning dataset to guide its learning of medical concepts; it then undergoes SVR, where reinforcement learning (RL) is applied to optimize spatial-textual alignment using Intersection-over-Union (IoU)-based rewards, ensuring that the generated phrases correspond to pixel-level anatomical evidence. In the Downstream Adapter Phase, BoxMed-RL's weights are frozen and a lightweight adapter is fine-tuned on raw reports, producing fluent, clinically coherent narratives while preserving the reasoning and spatial alignment learned earlier. The main contributions of our work are:

- We propose BoxMed-RL, a two-phase automated chest X-ray report generation model. To the best of our knowledge, BoxMed-RL is the first method to combine reasoning and spatial grounding in chest X-ray report generation task.
- We introduce medical concept learning to force models to internalize the diagnostic logic of radiologists through hierarchical reasoning steps, bridging the gap between free-form generation and clinically structured workflows.
- We introduce spatially verifiable reinforcement to explicitly align textual descriptions with anatomical regions using IoU-based rewards.
- Extensive experiments on 2 public datasets demonstrate that BoxMed-RL achieves an average 7 % improvement in natural language generation metrics compared to state-of-the-art methods. An average 5 % improvement in large language model-based metrics further underscores its robustness in generating high-quality chest X-ray reports.

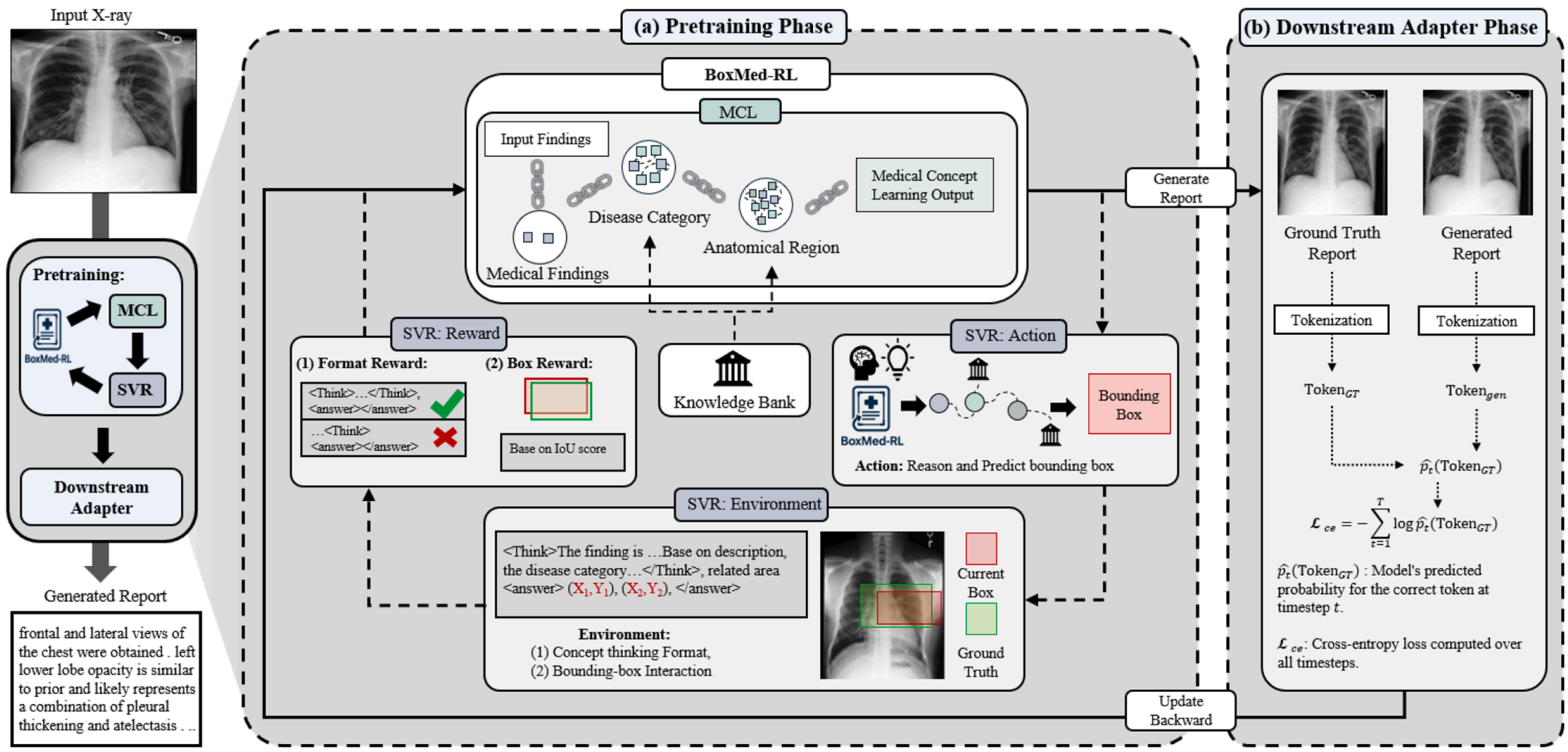

**Fig. 3.** Overview of the BoxMed-RL framework. (a) Pretraining phase integrates Medical Concept Learning (MCL) and Spatially Verifiable Reinforcement (SVR) to teach structured reasoning and spatial alignment. (b) Downstream adapter phase fine-tunes a lightweight adapter for fluent report generation while keeping pretrained weights frozen.

## 2. Related works

### 2.1. Automated medical report generation

With the advancements of deep learning in computer vision and natural language processing, many methods based on an encoder-decoder paradigm were proposed for automated generation of medical reports (Zhang et al., 2024c; Yang et al., 2022; Zhang et al., 2024b; Liu et al., 2021a,b). For the encoder component, these approaches differ in their input modalities. Several methods rely solely on X-ray images for report generation (Liu et al., 2021b).The first method (Liu et al., 2021b) leverages differences between diseased and normal X-ray images, while another method enhances the language loss function by incorporating more fine-grained word-level features(Chen et al., 2020). Other major approaches would incorporate cross-modal features, integrating text and images, to deliver additional expert insights, thus improving and directing the entire report generation process (Zhang et al., 2024c; Yang et al., 2022; Liu et al., 2021a). In particular, instead of relying on manually designed graphs, some methods let models autonomously learn knowledge graphs through node or edge prediction, thereby enhancing the learning process (Zhang et al., 2024c,b).

Beyond these approaches, recent works have further advanced medical report generation by incorporating structured reasoning or RL. One line of work introduced a CoT-like framework that reduces hallucinations by supervising intermediate reasoning steps, but without RL (Jiang et al., 2025). In contrast, RL has been applied in several ways: some methods directly optimize clinical quality to align reports with radiological standards (Zhou et al., 2024), others leverage AI feedback to adapt large language models for chest CT reporting (Yang et al., 2025a), while some focus on improving report fluency (Qin and Song, 2022). Together, these studies demonstrate diverse strategies to improve reliability and quality in report generation, based on reasoning supervision or optimization via RL.

### 2.2. Large vision-language models in report generation

Pretrained large language models, such as LLaMA (Touvron et al., 2023), GPT-4 (Achiam et al., 2023) have achieved remarkable success across a wide range of natural language processing tasks. However, due to the limited capability of large language models in handling cross-modal tasks, large vision-language models have been further developed to address multi-modality challenges by integrating both image and text inputs (Liu et al., 2024a; Wang et al., 2024; Chen et al., 2024). Emerging research refines the application of large vision-language models within the medical domain, tailoring them to diverse downstream tasks such as medical report generation, visual question-answering, and the proposal of treatment options based on provided modalities (Li et al., 2023; Pan et al., 2025). In contrast to the conventional encoder-decoder framework, which often requires task-specific customization of architecture and training methodologies, large vision-language models offer a more versatile approach. Pretrained or fine-tuned on extensive datasets comprising medical image-text pairs, large vision-language models demonstrate the capability to address a diverse range of tasks, including medical visual question answering, cross-modal report generation, and bounding box prediction, with greater adaptability and efficiency (Pan et al., 2025; Achiam et al., 2023; Wang et al., 2024).

### 2.3. RL in large vision-language models

After pretraining, large vision-language models require fine-tuning to adapt effectively to specific domains (Zhang et al., 2024a; Shengyu et al., 2023). While large vision-language models offer remarkable flexibility across various input modalities, their performance may lag behind specialized deep learning models. However, leveraging their pretrained knowledge and applying sufficient downstream fine-tuning, large vision-language models have the potential to outperform traditional deep learning models. Supervised fine-tuning and reinforcement fine-tuning are two primary approaches to adopt a model to specific tasks. Supervised fine-tuning leverages paired data, and optimizes with token-level losses (Zhang et al., 2024a; Shengyu et al., 2023). Reinforcement fine-tuning instead relies on reward signals, often derived from human or rule-based feedback, to guide learning (Ouyang et al., 2022; Chu et al., 2025). In medical report generation, RL from verifiable rewards addresses explainability by requiring models to justify predictions against predefined criteria. (Schulman et al., 2017; Guo et al., 2025; Chu et al., 2025). Specifically, rewards can be defined through verification functions, and optimized using algorithms such as proximal policy optimization or group relative policy optimization. This approach refines the model's decision-making process and enhances its explainability (Guo et al., 2025; Liu et al., 2025).

## 3. Method

### 3.1. Overview

We propose BoxMed-RL, a two-phase training framework designed to enhance the explainability and clinical reliability of automated chest X-ray report generation. Although BoxMed-RL leverages the Qwen2-VL-2B architecture (Wang et al., 2024) as the base vision–language model, the framework itself is model-agnostic. The two core components Medical Concept Learning (MCL) and Spatially Verifiable Reinforcement (SVR) can be integrated into any trainable large vision–language model. As illustrated in Fig. 3, the framework proceeds in two phases:

**Pretraining Phase.** BoxMed-RL first undergoes MCL, where we construct a structured dataset by decomposing medical findings into hierarchical reasoning steps (*Medical Findings → Disease Category → Anatomical Region*). This structured CoT dataset is used to guide training, enabling the model to internalize radiologists' diagnostic logic. It is then optimized with SVR, which applies RL to align textual descriptions with bounding box evidence via IoU-based rewards, ensuring verifiable spatial grounding. **Downstream Adapter Phase.** The pretrained weights are frozen, and a lightweight adapter is fine-tuned on raw reports. This step refines linguistic fluency and clinical phrasing while preserving the structured reasoning and alignment learned during pretraining.

In the BoxMed-RL framework, at Pretraining Phase (Section 3.2), the MCL (Section 3.2.1) and SVR (Section 3.2.2) are introduced to enhance its explainability through hierarchical reasoning and anatomical grounding, while Downstream Adapter Phase (Section 3.3) serves as the final step to ensure the generation of fluent and coherent report.

### 3.2. Pretraining phase

In this phase, BoxMed-RL is fine-tuned through MCL and SVR. These two modules equip BoxMed-RL with clinical knowledge and textual-

---

**Algorithm 1** Training Process of BoxMed-RL.

**Require:** X-ray Image $I$, CoT Annotations $C$, Prompt $q$, Bounding Boxes $B_{\text{gt}}$, Format $t$, Raw Report $O$
**Ensure:** BoxMed-RL ($\pi$) with reasoning and spatial grounding ability; Adapter ($\pi_{\text{Adapter}}$) for fluent report generation

1: **Phase I: Pretraining**
2: *Step 1: Medical Concept Learning (MCL)*
3: BoxMed-RL($\pi$) with initial parameters $\theta$
4: **for** $epoch = 0 \rightarrow epochs$ **do**
5: Given X-ray Image $I$, Annotations $C$
6: Compute MCL loss: $\mathcal{L}_{\text{MCL}}(\theta) \leftarrow (I, C, \pi_{\theta})$
7: Update parameters: $\theta' \leftarrow \arg\min \mathcal{L}_{\text{MCL}}(\theta)$
8: **end for**
// $\theta'$ is the updated parameters after MCL
9: *Step 2: Spatially Verifiable Reinforcement (SVR)*
10: **for** $epoch = 0 \rightarrow epochs$ **do**
11: Given prompt $q$, sample $N$ responses
12: $\{y_1, \dots, y_N\} \leftarrow (q, \pi_{\theta'})$
13: **for all** $y_i$ **do**
14: Compute IoU reward: $\mathcal{R}_{\text{IoU}}(q, y_i) \leftarrow \text{IoU}(q, y_i, B_{gt})$
15: Reward Combination: $R(q, y_i) \leftarrow \mathcal{R}_{\text{IoU}}(q, y_i) + \mathcal{R}_{\text{format}}(q, y_i, t)$
16: Compute SVR Expectation: $\mathbb{E}_{\text{SVR}}(\theta') \leftarrow R(q, y_i)$
17: Update parameters: $\theta'' \leftarrow \arg\max \mathbb{E}_{\text{SVR}}(\theta')$
18: **end for**
19: **end for**
// $\theta''$ is parameters after Pretraining
20: **Phase II: Downstream Adapter**
21: Freeze $\pi_{\theta''}$ and initialize Adapter $\pi_{\text{Adapter}}$ with parameters$\theta''$
22: **for** $epoch = 0 \rightarrow epochs$ **do**
23: Given X-ray Images $I$, Raw Reports $O$
24: Compute Adapter loss: $\mathcal{L}_{Aapter}(\theta'') \leftarrow (I, O, \pi_{Adapter(\theta'')})$
25: Update parameters: $\hat{\theta} \leftarrow \arg\min \mathcal{L}_{Adapter}(\theta'')$
26: **end for**
// $Adapter$'s parameters updated as $\hat{\theta}$
27: **return** BoxMed-RL($\pi$) with parameters $\theta''$ and Adapter($\pi_{\text{Adapter}}$) with parameters $\hat{\theta}$

---

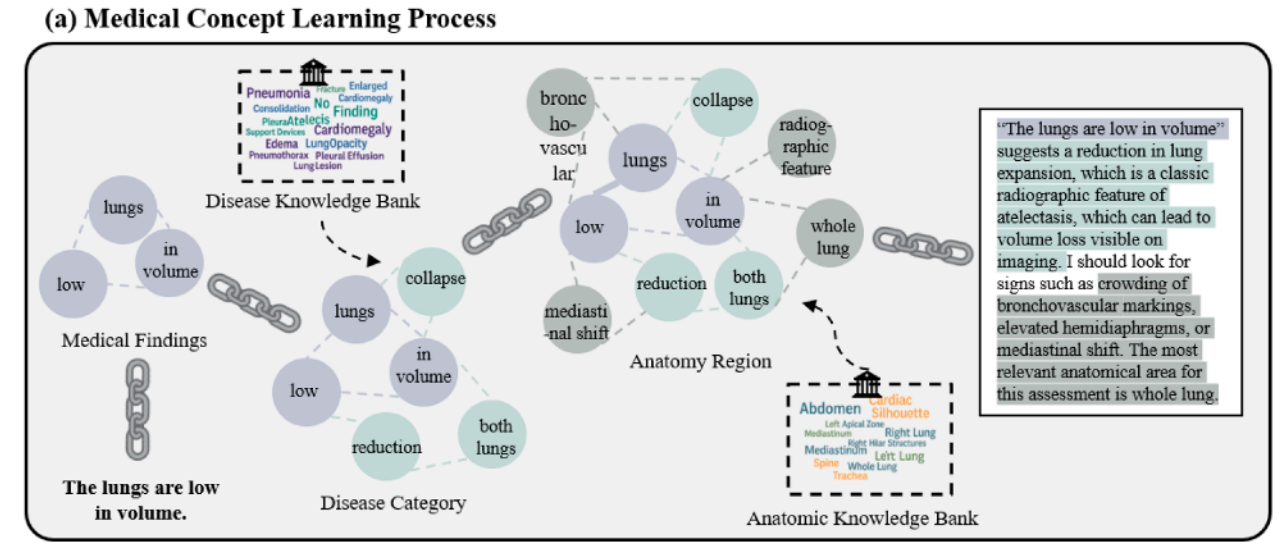

**(b) Prompt for MCL**

**Question:** Suppose you are a radiologist. Based on the description provided within <dcp>…</dcp> tags and related X-ray image, identify a suitable disease category from *disease category knowledge bank* and explicitly indicate the anatomical area(s) you inspected from *anatomical region knowledge bank*.

**Important:** Even if the category selected is No Finding, explicitly specify the anatomical area(s) inspected. Limit your choice of anatomical areas to the single most relevant one. If necessary for clarity, you may include one additional area (maximum two areas total). Do NOT list multiple sub-areas if a broader area like 'whole lung' adequately covers them.

**Format:** Your answer should strictly follow the format below:<think> Your reasoning process here... </think> <cat>Your chosen disease category here</cat> <area>Your chosen anatomical area(s) here (maximum two)</area>

**Your Answer:**

**Fig. 4.** Implementation details of the MCL module. (a) Example of CoT dataset construction: a raw finding is mapped to a disease category and localized to an anatomy region using the knowledge banks, forming a reasoning chain. (b) Prompt template used during MCL training.

**Table 1**
Comparison of different methods on MIMIC-CXR. BLEU evaluates phrase-level fluency; METEOR captures semantic similarity; ROUGE-L measures content coverage and structural overlap; CheXbert measures medical concept coverage; GEMA-Score evaluates both clinical accuracy and linguistic quality; and GreenScore captures clinically significant errors and matched findings. Results reproduced in our setting are marked with $^{\dagger}$. Statistical significance against these baselines is tested with a one-tailed Wilcoxon signed-rank test, with $^{*}$ indicating a p-value less than 0.05 and $^{**}$ indicating a p-value less than 0.01. Best scores are in **bold**.

| Method | Year | BLEU-1 | BLEU-2 | BLEU-3 | BLEU-4 | METEOR | ROUGE-L | GEMA-Score | CheXbert | GreenScore |
|---|---|---|---|---|---|---|---|---|---|---|
| Traditional Encoder-Decoder | | | | | | | | | | |
| R2Gen (Chen et al., 2020)$^{\dagger}$ | 2020 | 0.368$^{**}$ | 0.220$^{**}$ | 0.145$^{**}$ | 0.103$^{**}$ | 0.149$^{**}$ | 0.272$^{**}$ | 0.534$^{**}$ | 0.331$^{**}$ | 0.276$^{**}$ |
| GSK (Yang et al., 2022) | 2022 | 0.363 | 0.228 | 0.156 | 0.115 | - | 0.284 | - | 0.371 | - |
| RECAP (Hou et al., 2023) | 2023 | **0.429** | 0.267 | 0.177 | 0.125 | 0.168 | 0.288 | - | 0.393 | - |
| Region-Enhanced Encoder-Decoder | | | | | | | | | | |
| CMN (Chen et al., 2021)$^{\dagger}$ | 2021 | 0.353$^{**}$ | 0.213$^{**}$ | 0.140$^{**}$ | 0.098$^{**}$ | 0.162$^{**}$ | 0.279$^{**}$ | 0.530$^{**}$ | 0.281$^{**}$ | 0.295$^{**}$ |
| CMM (Qin and Song, 2022) | 2022 | 0.381 | 0.232 | 0.155 | 0.109 | 0.151 | 0.287 | - | 0.292 | - |
| Clinical-BERT (Yan and Pei, 2022) | 2022 | 0.383 | 0.230 | 0.151 | 0.106 | 0.144 | 0.275 | - | 0.415 | - |
| MA (Wang et al., 2022a) | 2022 | 0.396 | 0.244 | 0.162 | 0.115 | 0.151 | 0.274 | - | 0.389 | - |
| METransformer (Wang et al., 2023a) | 2023 | 0.386 | 0.250 | 0.169 | 0.124 | 0.152 | 0.291 | - | 0.311 | - |
| FMVP (Liu et al., 2024b) | 2024 | 0.389 | 0.236 | 0.156 | 0.108 | 0.150 | 0.284 | - | 0.336 | - |
| PromptMRG (Jin et al., 2024) | 2024 | 0.398 | - | - | 0.112 | 0.157 | 0.268 | - | 0.476 | - |
| RAMT (Zhang et al., 2024b) | 2024 | 0.362 | 0.229 | 0.157 | 0.113 | 0.153 | 0.284 | - | 0.335 | - |
| Prompts-Driven Large Language Models | | | | | | | | | | |
| RGRG (Tanida et al., 2023)$^{\dagger}$ | 2023 | 0.300$^{**}$ | 0.182$^{**}$ | 0.120$^{**}$ | 0.082$^{**}$ | 0.156$^{**}$ | 0.249$^{**}$ | 0.553$^{**}$ | 0.360$^{**}$ | 0.246$^{**}$ |
| R2GenGPT (Wang et al., 2023b) | 2023 | 0.409 | 0.261 | 0.179 | 0.128 | 0.160 | 0.287 | - | 0.428 | - |
| LM-RRG (Zhou et al., 2024) | 2024 | - | - | - | 0.122 | 0.165 | 0.296 | - | **0.484** | - |
| STREAM (Yang et al., 2025b) | 2025 | 0.420 | 0.267 | 0.184 | 0.133 | 0.164 | 0.291 | - | 0.462 | - |
| BoxMed-RL | | | | | | | | | | |
| Our BoxMed-RL | 2025 | 0.426 | **0.271** | **0.185** | **0.134** | **0.180** | **0.314** | **0.560** | 0.466 | **0.310** |

visual grounding, serving as a foundation for later phase to generate more trustworthy reports.

*3.2.1. Medical concept learning for radiologist-like structured reasoning*

The primary goal of the medical concept learning (MCL) module is to internalize the hierarchical diagnostic logic of radiologists, enabling transparent clinical reasoning. Inspired by the Chain-of-Thought prompting paradigm (Wei et al., 2022), we construct a structured radiological reasoning dataset that decomposes medical findings into hierarchical reasoning steps with three domain-specific conceptual levels:

*Findings → Disease Category → Anatomical Region*

This formulation reflects how radiologists review a case: they first identify a key finding (disease description or medical observation), then associate it with potential or relevant disease category, and finally localize it to the corresponding anatomical region for verification.

To design these anchors, we sought the advice of a senior radiologist. For the Disease Knowledge Bank, we adopt the 14 standardized CheXbert labels (Smit et al., 2020). For the Anatomic Knowledge Bank, we follow the 11 areas defined in STREAM (Yang et al., 2025b) and extend them with an additional "whole lung" region, based on the radiologist's suggestion that some findings (e.g., low lung volume, diffuse edema) span both lungs and cannot be anchored to a single side. Fig. 4(a) illustrates how a medical finding is converted into a CoT example: a finding is mapped to a disease category and then localized to an anatomy region, resulting in a reasoning chain (e.g., *lungs are low in volume → probable Atelectasis → suggests observing the Whole Lung*). The final MCL dataset comprises 3772 structured reasoning chains covering 14 disease categories and 12 anatomical regions. Detailed coverage for both disease categories and anatomical regions is provided in the supplementary material (see Fig. S1).

During MCL training, the model is supervised to reproduce such chains in a standardized format. Fig. 4(b) shows the prompt template used for training, the output is restricted to three stages: (i) an explanatory trace within `<think>` tags, where the model explains its interpretation of the finding; (ii) the most relevant or potential disease label within `<cat>` chosen from the Disease Knowledge Bank; and (iii) the most relevant anatomical region(s) within `<area>` from the Anatomical Region Knowledge Bank. At inference, these MCL outputs function solely as internal reasoning scaffolds and are not included in the clinical report output. The full reasoning trace can be optionally visualized for interpretability. Representative examples are provided in the supplementary material (Fig. S2). By requiring the model to articulate its reasoning before producing the final disease and anatomy predictions, the prompt mimics the workflow of radiologists and prevents the shortcut to an answer without justification. For a given training example consisting of a radiological image $I$ and structured CoT annotations $C$, the BoxMed-RL $\pi$'s parameters $\theta$ are optimized to $\theta'$ using a cross-entropy loss:

$$
\begin{aligned}
\theta' &= \arg\min_{\theta} \; \mathcal{L}_{\mathrm{MCL}}(\theta) \\
&= \arg\min_{\theta} \; -\sum_{t=1}^{T} \log p_{\theta}(c_t \mid I_t, c_{<t}),
\end{aligned} \tag{1}
$$

where $\theta'$ is the optimized set of model parameters after training with the MCL objective, $c_t$ denotes the $t^{th}$ token of the structured CoT annotation and $c_{<t}$ represents the preceding tokens.

By explicitly training the model to follow structured chains, BoxMed-RL learns to internalize diagnostic reasoning.

*3.2.2. Spatially verifiable reinforcement for anatomy location perception*

**Reinforcement Problem Statement:** While MCL module enables the model to mimic the flow of diagnostic reasoning, it lacks mechanisms to ensure spatial grounding of related textual descriptions. To bridge the gap between textual descriptions and anatomical evidence, we introduce the spatially verifiable reinforcement (SVR) module. Building on the MCL module, SVR employs RL with verifiable rewards through a group relative policy optimization framework to optimize two critical objectives: (1) Spatial consistency, ensuring that medical findings are aligned correctly with related anatomy regions. (2) Structured formatting, enforce the reasoning process before output the coordinate answer.

The overall process can be summarized as follows. In RL with verifiable rewards, given a query $q$, the policy model BoxMed-RL $\pi_{\theta'}$ (optimized after MCL) generates a response $y$, and receives the reward $R(q, y)$. The RL with verifiable rewards objective encourages high-reward generations while regularizing the policy via KL divergence against a reference model $\pi_{ref}$ (the frozen BoxMed-RL $\pi_{\theta'}$):

$$\begin{aligned}\theta'' &= \arg\max_{\theta'} \ \mathbb{E}_{\mathrm{SVR}}(\theta') \\ &= \arg\max_{\theta'} \mathbb{E}_{y\sim\pi_{\theta'}(q)}\left[R(q, y) - \beta \cdot \mathrm{KL}\left(\pi_{\theta'}(y|q) \,\|\, \pi_{\mathrm{ref}}(y|q)\right)\right],\end{aligned} \tag{2}$$

where $\theta''$ is the optimized set of model parameters after the SVR module and $\beta$ is the hyperparameter to control KL divergence.

**Table 2**
Comparison of different methods across various evaluation metrics in IU X-Ray, grouped by method type. The best results are highlighted in **bold**.

| Method | Year | BLEU-1 | BLEU-2 | BLEU-3 | BLEU-4 | METEOR | ROUGE-L | CheXbert |
|---|---|---|---|---|---|---|---|---|
| Traditional Encoder-Decoder | | | | | | | | |
| R2Gen (Chen et al., 2020) | 2020 | 0.475 | 0.304 | 0.219 | 0.170 | 0.187 | 0.371 | 0.530 |
| GSK (Yang et al., 2022) | 2022 | 0.496 | 0.327 | 0.238 | 0.178 | - | 0.381 | - |
| Region-Enhanced Encoder-Decoder | | | | | | | | |
| CMN (Chen et al., 2021) | 2021 | 0.475 | 0.309 | 0.222 | 0.170 | 0.191 | 0.375 | 0.552 |
| CMM (Qin and Song, 2022) | 2022 | 0.494 | 0.321 | 0.235 | 0.181 | 0.201 | 0.384 | - |
| Clinical-BERT (Yan and Pei, 2022) | 2022 | 0.495 | 0.330 | 0.231 | 0.170 | 0.209 | 0.376 | - |
| MA (Wang et al., 2022a) | 2022 | **0.501** | 0.328 | 0.230 | 0.170 | 0.213 | 0.386 | - |
| METransformer (Wang et al., 2023a) | 2023 | 0.483 | 0.332 | 0.228 | 0.172 | 0.192 | 0.380 | - |
| FMVP (Liu et al., 2024b) | 2024 | 0.485 | 0.315 | 0.225 | 0.169 | 0.201 | 0.398 | - |
| PromptMRG (Jin et al., 2024) | 2024 | 0.401 | - | - | 0.098 | 0.160 | 0.281 | 0.556 |
| RAMT (Zhang et al., 2024b) | 2024 | 0.482 | 0.310 | 0.221 | 0.165 | 0.195 | 0.377 | - |
| Prompts-Driven Large Language Models | | | | | | | | |
| R2GenGPT (Wang et al., 2023b) | 2023 | 0.491 | 0.323 | 0.234 | 0.180 | 0.211 | 0.376 | 0.601 |
| LM-RRG (Zhou et al., 2024) | 2024 | - | - | - | **0.208** | 0.216 | 0.387 | - |
| STREAM (Yang et al., 2025b) | 2025 | 0.499 | **0.333** | 0.238 | 0.178 | 0.213 | 0.377 | 0.608 |
| BoxMed-RL | | | | | | | | |
| Our BoxMed-RL | 2025 | 0.490 | 0.320 | **0.240** | 0.178 | **0.220** | **0.404** | **0.610** |

**Table 3**
Classification metrics across 14 categories of CheXbert.

| Category | Accuracy | Precision | Recall | F1-Score |
|---|---|---|---|---|
| No Finding | 0.892 | 0.651 | 0.739 | 0.681 |
| Cardiomediastinum | 0.594 | 0.389 | 0.386 | 0.386 |
| Cardiomegaly | 0.581 | 0.418 | 0.419 | 0.418 |
| Lung Lesion | 0.899 | 0.361 | 0.341 | 0.338 |
| Lung Opacity | 0.579 | 0.397 | 0.393 | 0.388 |
| Edema | 0.689 | 0.429 | 0.422 | 0.425 |
| Consolidation | 0.838 | 0.380 | 0.363 | 0.368 |
| Pneumonia | 0.756 | 0.383 | 0.369 | 0.373 |
| Atelectasis | 0.623 | 0.418 | 0.424 | 0.420 |
| Pneumothorax | 0.927 | 0.372 | 0.378 | 0.375 |
| Pleural Effusion | 0.648 | 0.457 | 0.457 | 0.456 |
| Pleural Other | 0.937 | 0.319 | 0.326 | 0.323 |
| Fracture | 0.911 | 0.335 | 0.334 | 0.331 |
| Support Devices | 0.739 | 0.485 | 0.492 | 0.488 |

To improve sample efficiency and avoid the need for a separate value estimator, we adopt the group relative policy optimization algorithm (Guo et al., 2025). Unlike standard proximal policy optimization (Schulman et al., 2017), which requires explicit estimation of the value function and constrains policy updates through a clipped surrogate objective, group relative policy optimization compares multiple responses within a group and normalizes rewards relatively. This relative comparison reduces variance and stabilizes the training without an additional value network. Such efficiency makes group relative policy optimization particularly well suited to our setting, where reward signals (IoU-based alignment and structured format verification) are sparse and discrete. The algorithm operates by generating a group of responses for each query $q$:

$$\{y_1, \ldots, y_N\} \sim \pi_{\theta'}(y \mid q), \tag{3}$$

where $N$ is the number of groups.

For each response $y_i$, compute the total verifiable reward $r_i$, then compute the corresponding group relative policy optimization normalized score $a_i$:

$$a_i = \frac{r_i - \mathrm{mean}(\{r_1, \ldots, r_N\})}{\mathrm{std}(\{r_1, \ldots, r_N\})}. \tag{4}$$

We then substitute these scores into the RL with verifiable rewards framework in Eq. (2), resulting in the following hybrid optimization objective:

$$\theta'' = \arg\max_{\pi_{\theta'}} \ \mathbb{E}_{y_i\sim\pi_{\theta'}(q)}\left[a_i - \beta \cdot \mathrm{KL}\left(\pi_{\theta'}(y_i|q) \,\|\, \pi_{\mathrm{ref}}(y_i|q)\right)\right], \tag{5}$$

where $a_i$ reflects the relative advantage of response $y_i$ within the group. This hybrid strategy allows BoxMed-RL to explicit supervision of verifiable reward signals with the efficiency and stability of relative policy optimization.

**IoU Reward for Spatially Alignment:** To promote anatomically precise and spatially verifiable ability, we incorporate an IoU-based reward as a core component of our RL strategy. This reward function explicitly guides the model to align diagnostic descriptions with correct image regions, reinforcing textual-visual grounding during policy optimization.

Each predicted bounding box is represented by its corner coordinates $(x_1, y_1)$ and $(x_2, y_2)$, where $(x_1, y_1)$ defines the top-left corner and $(x_2, y_2)$ defines the bottom-right corner of the bounding area. The IoU between a predicted box $B_{pred}$ and a ground truth bounding box $B_{gt}$ is computed as:

$$\mathrm{IoU}(B_{\mathrm{pred}}, B_{\mathrm{gt}}) = \frac{|B_{\mathrm{pred}} \cap B_{\mathrm{gt}}|}{|B_{\mathrm{pred}} \cup B_{\mathrm{gt}}|}, \tag{6}$$

where $\mathrm{IoU} \in [0, 1]$.

This metric quantifies the degree of overlap between the predicted and actual regions. A high IoU score indicates strong spatial alignment, while a low score penalizes incorrect or irrelevant location. We then define the spatial reward as:

$$\mathcal{R}_{IoU} = \begin{cases} \mathrm{IoU}, & \text{if IoU} > 0.3 \\ 0, & \text{otherwise} \end{cases}. \tag{7}$$

This reward is integrated into our total verifiable reward $R$, alongside a format reward $\mathcal{R}_{format}$ that ensures structural correctness of the generated output.

**Table 4**
Ablation study results on different components of BoxMed-RL. Results are reported as mean±std over multiple runs. A one-tailed Wilcoxon signed-rank test is performed on each ablated model, with * indicating a p-value less than 0.05 and ** indicating a p-value less than 0.01.

| Configuration | MCL | SVR | Frozen | Adapter | BLEU-1 | BLEU-2 | BLEU-3 | BLEU-4 | METEOR | ROUGE-L | GEMA-Score | CheXbert | GreenScore |
|---|---|---|---|---|---|---|---|---|---|---|---|---|---|
| w/o MCL & SVR | ✗ | ✗ | ✓ | ✓ | 0.385±0.113** | 0.235±0.108** | 0.156±0.106** | 0.110±0.101** | 0.173±0.063** | 0.285±0.100** | 0.531±0.085** | 0.393±0.178** | 0.270±0.202** |
| w/o SVR | ✓ | ✗ | ✓ | ✓ | 0.393±0.105** | 0.239±0.104** | 0.159±0.105** | 0.112±0.101** | 0.173±0.061** | 0.288±0.097** | 0.540±0.074** | 0.401±0.171** | 0.280±0.208** |
| w/o MCL | ✗ | ✓ | ✓ | ✓ | 0.392±0.102** | 0.241±0.100** | 0.160±0.101** | 0.112±0.099** | 0.177±0.062** | 0.290±0.098** | 0.551±0.080** | 0.422±0.191** | 0.292±0.201** |
| w/o Frozen | ✓ | ✓ | ✗ | ✗ | 0.435±0.093 | 0.276±0.101 | 0.173±0.098** | 0.129±0.095** | 0.174±0.064** | 0.298±0.101** | 0.550±0.078** | 0.457±0.204** | 0.289±0.211** |
| BoxMed-RL | ✓ | ✓ | ✓ | ✓ | 0.426±0.098 | 0.271±0.104 | 0.185±0.109 | 0.134±0.111 | 0.180±0.066 | 0.314±0.107 | 0.560±0.082 | 0.466±0.213 | 0.310±0.200 |

By leveraging IoU as a direct spatial alignment metric, our approach encourages BoxMed-RL to remain textual-visual consistency.

**Format Reward for Structured Output:** To complement spatial alignment, we enforce the output format using a binary reward signal $\mathcal{R}_{\text{format}}$:

$$\mathcal{R}_{\text{format}} = \begin{cases} 1, & \text{if } t_y \in t \\ 0, & \text{otherwise} \end{cases}, \tag{8}$$

where $t_y$ is the response $y$'s format, $t$ has the form of $< think > \ldots < /think >, < answer > \ldots < /answer >$. This format reward encourages the model to output its reasoning process and spatial answer in a structured format.

**Reward Combination and Objective Updated:** Given the query $q$, the total verifiable reward $R$ for each response $y_i$ is the additive combination of $\mathcal{R}_{\text{IoU}}$ and $\mathcal{R}_{\text{format}}$:

$$R(q, y_i) = \mathcal{R}_{\text{IoU}}(q, y_i) + \mathcal{R}_{\text{format}}(q, y_i). \tag{9}$$

We then use Eq. (9) to substitute $r_i$ in Eq. (4), thus Eq. (4) is updated as:

$$A(q, y_i) = \frac{R(q, y_i) - \text{mean}\left(\{R(q, y_j)\}_{j=1}^{N}\right)}{\text{std}\left(\{R(q, y_j)\}_{j=1}^{N}\right)}. \tag{10}$$

Therefore, our hybrid optimization objective in Eq. (5) can be updated as:

$$\theta'' = \arg\max_{\pi'_\theta} \mathbb{E}_{y_i \sim \pi_{\theta'}(q)}\left[A(q, y_i) - \beta \cdot \text{KL}\left(\pi_{\theta'}(y_i|q) \parallel \pi_{\text{ref}}(y_i|q)\right)\right]. \tag{11}$$

This formulation jointly optimizes for verifiable structure and anatomical alignment, allowing BoxMed-RL to correctly relate descriptions with spatial information.

### 3.3. Downstream adapter phase

Following Pretraining, we freeze the BoxMed-RL backbone and attach a lightweight LoRA (Hu et al., 2022) adapter $\pi_{Adapter}$ to refine the fluency and clinical fidelity of the generated reports. While Pretraining emphasizes structured diagnostic reasoning and spatial grounding, the purpose of this adapter is not to overwrite these learned capabilities but rather to enhance the surface-level linguistic quality by aligning the outputs to the distribution of professional radiology reports.

Specifically, we train $Adapter$, initialized from the pretrained parameters $\theta''$, using a large corpus of chest X-ray images $\{I_1, \ldots, I_M\}$ paired with corresponding radiology reports $\{o_1, \ldots, o_M\}$. The $Adapter$'s parameters $\hat{\theta}$ are optimized using a cross-entropy loss:

$$\begin{aligned} \hat{\theta} &= \arg\min_{\theta''} \ \mathcal{L}_{Adapter}(\theta'') \\ &= \arg\min_{\theta''} \ -\sum_{m=1}^{M} \log p_{\theta''}(y_m | I_m, y_{<m}), \end{aligned} \tag{12}$$

where $p_{\theta''}(y_m|I, y_{<m})$ denotes the probability of generating token $y_m$ given the input image $I_m$ and previously generated tokens $y_{<m}$. By updating only the adapter's weights, the pretrained reasoning and spatial-grounding skills remain preserved, enabling $Adapter$ to effectively learn expert clinical phrasing and linguistic fluency suitable for healthcare workflows. The learning process of our BoxMed-RL can be summarized as Algorithm 1.

## 4. Experiments and analysis

### 4.1. Datasets and evaluation metrics

In the Pretraining Phase of BoxMed-RL, we employ two bounding box–sentence pair datasets: MS-CXR (Boecking et al., 2022) and LATTE-CXR (Ghelichkhan and Tasdizen, 2025). MS-CXR, sourced from MIMIC-CXR (Johnson et al., 2019), includes 1162 image–sentence pairs with annotated bounding boxes and corresponding medical phrases. LATTE-CXR, built from MIMIC-CXR and REFLACX (Bigolin Lanfredi et al., 2022), offers 13,751 verified bounding box annotations aligned with radiological findings. Both datasets label samples with medical findings and their spatial regions. In the Downstream Adapter Phase, we use two radiology report generation benchmarks: MIMIC-CXR and IU X-Ray (Demner-Fushman et al., 2016). MIMIC-CXR comprises 377,110 chest X-ray images paired with 227,835 radiology reports, while IU X-Ray includes 7470 images and 3955 reports. For both, we target the “findings" section of each report for generation.

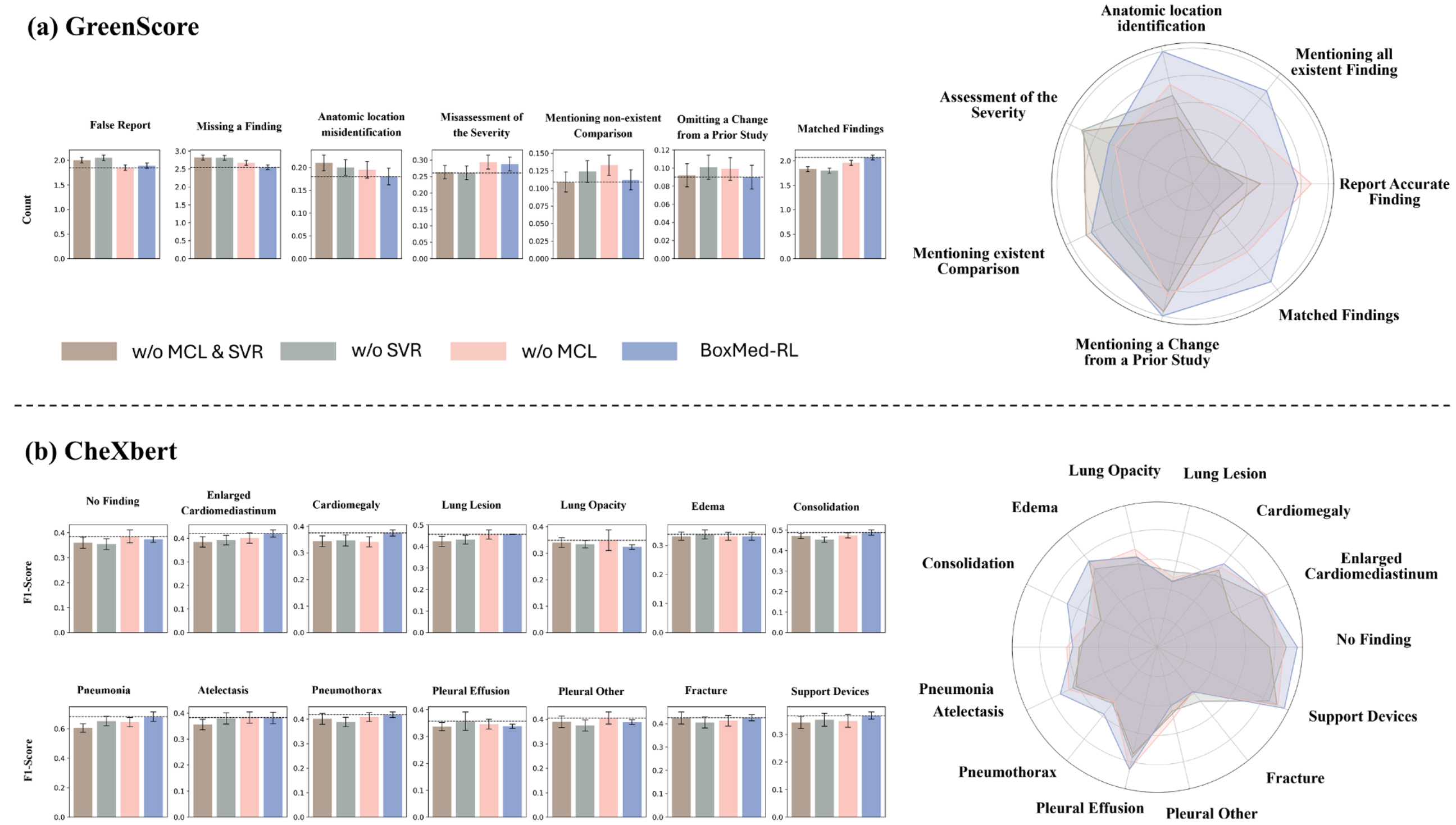

**Fig. 5.** Comprehensive analysis of ablation modules on GreenScore and CheXbert. Bar charts show detailed values for the 7 GreenScore clinical criteria categories and 14 CheXbert disease classes, with the black line indicating the best score in each. Error bars denote the standard deviation across runs. GreenScore captures clinically significant errors (lower is better) and matched findings (higher is better), while CheXbert assesses clinical accuracy (higher is better). The radar chart summarizes overall trends.

For external validation and robustness to scanner and patient mix shifts, we use a filtered CheXpert Plus subset (Chambon et al., 2024); curation and preprocessing details are in the supplementary material.

To ensure data integrity and avoid leakage from pretraining, we used the official MIMIC-CXR split for downstream tasks and verified split consistency across datasets. Since MS-CXR and LATTE-CXR are derived from MIMIC-CXR, we cross-matched DICOM identifiers and removed inconsistent assignments. This filtering ensures that no image used for pretraining appears in a different MIMIC-CXR split. Details of this verification and filtering are provided in the supplementary material.

We evaluate BoxMed-RL using three metric categories. First, to evaluate fluency, coherence, and lexical similarity, we employ standard natural language generation (NLG) metrics: BLEU (Papineni et al., 2002), ROUGE-L (Recall-oriented) (Lin, 2004), and METEOR (Lavie and Agarwal, 2007). Second, we measure clinical efficacy (CE) by evaluating how well the generated reports capture medically relevant concepts by using CheXbert (Smit et al., 2020). This is quantified using classification metrics and cosine similarity. Third, we adopt advanced large language model based metrics: GEMA-Score (Zhang et al., 2025) evaluates both clinical accuracy and linguistic quality, and GreenScore (Ostmeier et al., 2024) measures clinical reliability by counting matched findings and categorizing clinically significant errors.

### 4.2. Implementation details

All chest X-ray images are resized to one-quarter of their original height and width to preserve aspect ratio while reducing memory consumption. Correspondingly, all bounding box coordinates are resized using the same ratio to maintain consistency in spatial supervision. For text targets, we extract the ‘findings’ section from each radiology report, removing excess whitespace, special characters, and entries with empty or missing findings. In the Pretraining Phase, BoxMed-RL trains MCL and SVR modules sequentially using the AdamW optimizer. In the downstream adapter phase, the backbone is frozen while a lightweight LoRA adapter is trained. The detailed implementation details are provided in supplementary material.

### 4.3. Quantitative results

We conduct a comprehensive comparative study of various frameworks, including **traditional encoder-decoder models**: R2Gen (Chen et al., 2020), GSK (Yang et al., 2022), RECAP (Hou et al., 2023); **region-enhanced encoder-decoder models**: MA (Wang et al., 2022a), CMN (Chen et al., 2021), CMM (Qin and Song, 2022), Clinical-Bert (Yan and Pei, 2022), MEtransformer (Wang et al., 2023a), PromptMRG (Jin et al., 2024), FMVP (Liu et al., 2024b); and **prompt-driven large language models**: STREAM (Yang et al., 2025b), R2GenGPT (Wang et al., 2023b), RGRG (Tanida et al., 2023), LM-RRG (Zhou et al., 2024). Under the defined frameworks, the deep learning models employ distinct architectures for visual feature extraction and decoding. ResNet serves as the visual encoder in R2Gen, CMN, CMM, GSK, and PromptMRG. In contrast, DenseNet is utilized by Clinical-BERT, RAMT, FMVP, and MA to capture dense visual representations. RGRG adopts Faster R-CNN for region-specific feature extraction through bounding box detection. Meanwhile, Vision Transformers are employed by METransformer, R2GenGPT, RECAP, and STREAM, harnessing self-attention mechanisms to model global image context. For decoding, RGRG, R2GenGPT, LM-RRG, and STREAM integrate large language models to generate coherent reports, capitalizing on their advanced generative abilities. Conversely, the remaining frameworks utilize transformer-based architectures or their variants as decoders, enabling effective sequential text generation. BoxMed-RL differs from other models by employing a large vision-language model that merges visual and textual encoder functions into a single unified model.

**Table 5**
Comparison of different training strategies for the adaptation of the report generation task. Results are reported as mean±std over multiple runs. A one-tailed Wilcoxon signed-rank test is performed on each strategy, with * indicating a p-value less than 0.05 and ** indicating a p-value less than 0.01.

| Vision Tower | Projector | Adapter | BLEU-1 | BLEU-2 | BLEU-3 | BLEU-4 | METEOR | ROUGE-L | CheXbert |
|---|---|---|---|---|---|---|---|---|---|
| ✓ | ✗ | ✗ | 0.379±0.115$^{**}$ | 0.230±0.113$^{**}$ | 0.151±0.110$^{**}$ | 0.105±0.113$^{**}$ | 0.171±0.069$^{**}$ | 0.282±0.103$^{**}$ | 0.442±0.211$^{**}$ |
| ✗ | ✓ | ✗ | 0.380±0.113$^{**}$ | 0.234±0.111$^{**}$ | 0.155±0.123$^{**}$ | 0.109±0.120$^{**}$ | 0.175±0.071$^{**}$ | 0.287±0.106$^{**}$ | 0.447±0.203$^{**}$ |
| ✓ | ✗ | ✓ | 0.390±0.107$^{**}$ | 0.237±0.109$^{**}$ | 0.157±0.105$^{**}$ | 0.107±0.115$^{**}$ | 0.173±0.061$^{**}$ | 0.288±0.104$^{**}$ | 0.444±0.213$^{**}$ |
| ✗ | ✓ | ✓ | 0.394±0.101$^{**}$ | 0.242±0.110$^{**}$ | 0.161±0.121$^{**}$ | 0.114±0.110$^{**}$ | 0.174±0.065$^{**}$ | 0.288±0.098$^{**}$ | 0.443±0.209$^{**}$ |
| ✗ | ✗ | ✓ | 0.426±0.098 | 0.271±0.104 | 0.185±0.109 | 0.134±0.111 | 0.180±0.066 | 0.314±0.107 | 0.466±0.213 |

#### 4.3.1. Natural language generation metrics

BoxMed-RL demonstrates significant performance in NLG metrics, achieving state-of-the-art performance in BLEU-3 and BLEU-4 scores across two datasets: MIMIC-CXR (0.185 and 0.134, Table 1) and IU X-ray (0.240 and 0.178, Table 2). It also yields an average improvement of 7 % in both METEOR and ROUGE-L, metrics that emphasize semantic adequacy and content coverage. These consistent gains on external datasets not included in pretraining underscore the robustness and generalizability of our framework.

It is important to note that BoxMed-RL does not always achieve the highest BLEU-1 or BLEU-2 scores, which mainly reflect surface-level word overlap. This outcome is partly influenced by the MCL, where the model learns to think like a radiologist instead of memorizing individual lexical patterns. As a result, models such as RECAP and MA that emphasize fluency and lexical similarity may obtain slightly higher unigram or bigram matches, while BoxMed-RL produces clinically meaningful multi-word expressions and maintains structural coherence. Consequently, its advantages are more evident in BLEU-3/4, METEOR, and ROUGE-L, which better capture phrase-level fluency, synonym usage, and content coverage.

We also highlight that absolute BLEU values in radiology remain lower than in general text generation tasks, a well-documented characteristic of this domain due to long report length, high lexical variability, and the existence of multiple clinically valid phrasings (Chen et al., 2020; Jin et al., 2024; Tanida et al., 2023). Previous studies have reported similarly incremental gains despite substantial methodological advances, underscoring that relative improvements are more informative than absolute scores. In this context, BoxMed-RL's consistent gains in BLEU-3/4, METEOR, and ROUGE-L provide a more reliable indicator of progress. To complement these surface-level evaluations, we report clinical metrics that capture reliability in the following section.

By associating observations with relevant medical concepts, BoxMed-RL enhances contextual accuracy and appropriate terminology usage, thus improving higher-order BLEU and recall-oriented metrics. This balance between linguistic adequacy and clinical grounding differentiates BoxMed-RL from models optimized solely for surface fluency, aligning with our overall goal of verifiable and transparent report generation.

#### 4.3.2. Clinical efficacy metrics

BoxMed-RL demonstrates stable performance in analyzing clinical concepts, achieving a CheXbert F1 score of 0.466 on MIMIC-CXR and a cosine similarity of 0.610 on IU X-ray. This consistency contrasts with PromptMRG, which shows unstable performance, ranking second in MIMIC-CXR but dropping to fourth in IU X-ray.

As shown in Table 1, the top four CheXbert scores: LM-RRG (0.484), PromptMRG (0.476), BoxMed-RL (0.466), and STREAM (0.462) are clearly higher than the rest. Among these, BoxMed-RL is the only model not explicitly designed to optimize clinical efficacy metrics. STREAM leverages region-level semantic entity prompts retrieved from a cross-modal knowledge bank to enhance clinical concept alignment, while LM-RRG and PromptMRG directly incorporate clinical supervision signals, either through diagnosis-aware prompts or reinforcement learning rewards tied to clinical assessment. Such targeted designs naturally raise CheXbert scores but often come with trade-offs in lexical diversity and fluency, as reflected in their weaker NLG results. In contrast, BoxMed-RL attains competitive CheXbert performance without relying on metric-specific shortcuts, which explains both its slightly lower absolute score and its greater stability across datasets.

Further evidence of this balance is shown in Table 3, where BoxMed-RL achieves strong accuracy in detecting “No Finding” and “Support Devices” while maintaining consistent F1 scores across disease categories (ranging from 0.456 to 0.323). This relatively narrow range highlights its reliability across categories. Moreover, when assessed with large language model–based metrics of readability, coherence, and completeness, BoxMed-RL shows an average 5 % improvement, underscoring its robustness in producing clinically accurate yet linguistically sound reports. Overall, BoxMed-RL delivers balanced and stable performance in all three types of metrics, while uniquely enhancing explainability and verifiability.

#### 4.3.3. Different objectives of RL in report generation

To further contextualize our approach, we compare BoxMed-RL with recent RL-based methods for chest X-ray report generation. While all three frameworks adopt RL, they differ in both objectives and outcomes. CMM employs RL to enhance lexical similarity and sentence fluency. As a result, it achieves strong performance in traditional NLG metrics such as BLEU, METEOR, and ROUGE-L (see Table 1), often reaching near state-of-the-art among methods from 2022 and earlier. However, its improvements remain largely at the surface level of text similarity, without addressing clinical reasoning or anatomical grounding. LM-RRG directly incorporates RL through a clinical quality assessment reward. This leads to substantial gains in the coverage of medical concept, reflected in its highest CheXbert scores among all baselines. However, its reliance on clinical efficacy rewards results in weaker performance on NLG metrics, highlighting a trade-off between clinical alignment and linguistic fluency. BoxMed-RL, in contrast, does not apply RL to directly optimize for textual similarity or clinical quality scores. Instead, our framework leverages SVR, where RL enforces textual–visual alignment through bounding-box supervision. This ensures that generated descriptions are explicitly linked to anatomical regions, enhancing interpretability. Combined with MCL, BoxMed-RL uniquely integrates structured reasoning with spatial grounding. Consequently, our method achieves balanced improvements in both NLG and clinical metrics, not by optimizing a single evaluation criterion, but by producing reports that are more transparent and explainable.

In addition, to validate the effectiveness of the proposed RL strategy in SVR, we evaluated the spatial grounding ability of BoxMed-RL on the held-out MS-CXR test set. Quantitative localization analysis using mean Intersection-over-Union (mIoU) and mean Average Precision (mAP) demonstrates that BoxMed-RL achieves the highest grounding accuracy among comparison methods. These results confirm that the SVR module not only improves textual and clinical consistency but also enhances spatial verifiability by aligning generated findings with anatomically relevant regions. Detailed quantitative comparisons and statistical analyses are provided in the supplementary material (Fig. S3).

| | Generated/GT Matched Reports | BoxMed-RL's Correct Reasoning Process | Generated/GT Regions |
|---|---|---|---|
| (a) | **BoxMed-RL's Report:**<br>… there is mild cardiomegaly. subtle fluid along the right major fissure, probable mild pulmonary edema…<br>**Ground Truth Report:**<br>…cardiac silhouette is enlarged. mild peribronchial cuffing and trace fluid along the right major fissure, probable mild pulmonary edema... | **BoxMed-RL's Reasoning Process:**<br>**there is mild cardiomegaly**<br><think>The description related to the **heart size**. The most relevant disease category is **cardiomegaly**. The anatomical area inspected is the **cardiac silhouette** as it… <think><br>**subtle fluid… probable mild pulmonary edema**<br><think>The description suggests fluid… aligns …**pulmonary edema**… The right major fissure is part of the right lung, so the relevant anatomical area would be the right lung. </think> | |
| (b) | **BoxMed-RL's Report:**<br>left-sided port-a-cath tip…no focal consolidation, pleural effusion or pneumothorax is identified.<br>**Ground Truth Report:**<br>there is a left chest port with tip at the cavoatrial junction… no pleural effusion or pneumothorax… | **BoxMed-RL's Reasoning Process:**<br>**left-sided port-a-cath tip**<br><think> The description indicates **the presence of a support device**…The most relevant anatomical area for the placement of a chest port is the **mediastinum**.</think><br>**no focal consolidation, pleural effusion or pneumothorax**<br><think>The description indicates... the most appropriate category is **No Finding**. The relevant anatomical area inspected is the **whole lung**...</think> | |
| (c) | **BoxMed-RL's Report:**<br>the patient has received a left-sided chest tube …the hear size is normal…<br>**Ground Truth Report:**<br>there is a there is a left sided chest tube. … the heart and pulmonary vessels are normal… | **BoxMed-RL's Reasoning Process:**<br>**the patient has received a left-sided chest tube**<br><think> The description indicates the presence of **a medical intervention device**. This corresponds to **support devices**. The relevant anatomical area is the **left lung**..</think><br>**the hear size is normal**<br><think>The description excludes the presence of cardiomegaly. ... the disease category is **no finding**. The relevant anatomical area inspected is the **cardiac silhouette**....</think> | |
| (d) | **BoxMed-RL's Report:**<br>there is mild opacification in the right lung base …left basilar retrocardiac opacities are present with consolidation…<br>**Ground Truth Report:**<br>right lower lung opacity may be atelectasis… retrocardiac left basilar opacity also likely represents consolidation… | **BoxMed-RL's Reasoning Process:**<br>**there is mild opacification in the right lung base**<br><think> The description identifies **the opacification in right lung**… suggest category is **lung opacity**. The relevant anatomical area inspected is the **right lung**...</think><br>**left basilar retrocardiac opacities …with consolidation**<br><think> The description notes **consolidation**…the appropriate disease category is **consolidation**. The relevant anatomical is the **left lung**... think> | |

**Fig. 6.** Qualitative examples of BoxMed-RL's report generation, reasoning process and detected anatomy regions. Each row presents a matched generated/ground truth report, BoxMed-RL's interpretable reasoning chain, and predicted (dashed boxes)/ground truth anatomical regions (solid boxes). Report findings are color-coded according to their corresponding anatomical regions.

### 4.4. Ablation studies

#### 4.4.1. Frozen vs. non-Frozen pretraining weights

In BoxMed-RL, we freeze the pretrained backbone weights during the Downstream Adapter Phase to preserve the structured reasoning and spatial alignment acquired in the Pretraining Phase (MCL + SVR). This design allows the lightweight adapter to focus on report fluency without overriding internalized diagnostic logic.

To validate this choice, we compared BoxMed-RL with a non-frozen variant (denoted as *w/o Frozen* in Table 4), where all backbone parameters were updated by supervised fine-tuning. The non-frozen variant achieved slightly higher BLEU-1 and BLEU-2 scores, but consistently underperformed in higher-order BLEU scores, METEOR, ROUGE-L, CheXbert, and large language model–based metrics (GEMA-Score and GreenScore). These results suggest that full fine-tuning may improve surface-level n-gram overlap but destabilize the previous internalized ability. Moreover, qualitative inspection revealed that the non-frozen model frequently lost the ability to articulate its reasoning and spatial observations learned during the Pretraining Phase. In contrast, freezing the pretrained backbone stabilizes learning and leads to better overall performance across evaluation metrics, supporting our choice of adopting frozen weights with a task-specific adapter in the Downstream Phase.

#### 4.4.2. Impacts of the components in BoxMed-RL

To better highlight the individual contributions of MCL and SVR, we explicitly report ablations that remove each component separately (*w/o MCL* and *w/o SVR*) as well as both together. This allows us to isolate

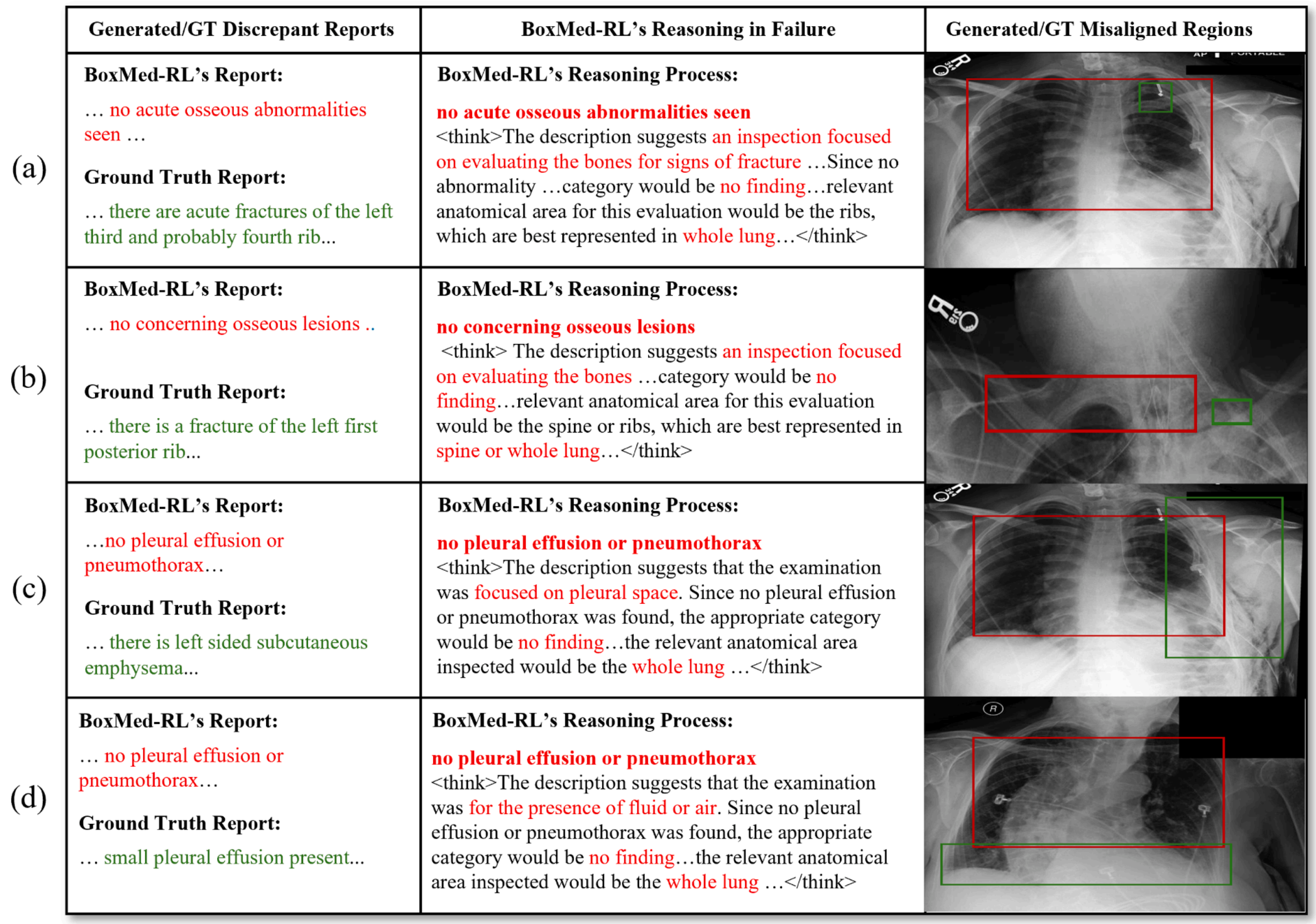

**Fig. 7.** Representative failure examples from BoxMed-RL. Each row shows the generated and reference reports, BoxMed-RL's reasoning chain, and predicted (red boxes) and ground truth (green boxes) anatomical regions. Although the final predictions are incorrect, the model demonstrates interpretable behavior by consistently inspecting relevant anatomical areas and offering transparent reasoning trace and diagnostic insight.

how structured reasoning and spatial alignment each contribute to the overall performance.

The results, presented in Table 4, demonstrates that the final framework, BoxMed-RL, leads to the best performance. With adapter alone, no extra knowledge is acquired through MCL or SVR. Consequently, the large vision-language model solely develops the capacity to generate reports from X-ray images, focusing on report generation proficiency. However, it lacks the ability to reasoning and make observations. For example, while it may detect an abnormality in an X-ray image, it might not identify the specific region or provide additional cues that could aid in further diagnostic analysis. Conversely, incorporating either MCL or SVR improves performance in NLG metrics, with average gains of 3 % (for SVR) and 7 % (for MCL) in BLEU scores, and disease classification metrics. Specifically, removing MCL reduces BLEU-3/4 and CheXbert, while removing SVR causes larger drops in GreenScore and GEMA-Score. This highlights that MCL primarily enhances structured reasoning and diagnostic coverage, whereas SVR strengthens spatial grounding and clinical consistency. These improvements stem from the integration of radiologist domain expertise and visual cues from the image. Without MCL, the model's performance is comparable to that without SVR but remains slightly lower (5 %) in disease classification metrics. This is because having the ability to observe X-ray images helps in identifying abnormalities. Although SVR aligns sentences with image features, enabling large vision-language models to better identify regions, it does not provide the reasoning capabilities derived from MCL.

By integrating all components, BoxMed-RL demonstrates superior X-ray report generation capabilities, as evidenced by an average 22 % improvement in the BLEU score and a 15 % increase in the CE metric. Its reasoning and visual alignment capabilities enable the model to effectively leverage text-image relationships.

As detailed in Fig. 5, BoxMed-RL demonstrates significant performance improvements compared to other ablated models. From the perspective of the GreenScore, we evaluated each model on seven clinical evaluation criteria: "False report of a finding in the candidate", "Missing a finding present in the reference", "Misidentification of a finding's anatomic location/position", "Misassessment of the severity of a finding", "Mentioning a comparison that is not in the reference", "Omitting a comparison detailing a change from a prior study", and "the number of matched findings" (Ostmeier et al., 2024). For the first six criteria, the Green Score counts errors, so lower values indicate better performance; for the number of matched findings, higher values are better. The radar chart shows that BoxMed-RL achieves holistic performance across all seven areas, with an average improvement of 6 % compared to other ablated models. Notably, for misidentification of anatomic location, BoxMed-RL and the model without MCL perform best, as they include the SVR module for visual alignment. The CheXbert 14 categories classification results are shown following the GreenScore. Among the 14 classes, BoxMed-RL demonstrates the best performance in eight classes, including Enlarged Cardiomediastinum, Cardiomegaly, Lung Lesion, Consolidation, Support Devices, Fracture, Pneumothorax, and Atelectasis, with an average percentage improvement of 5.11 % compared to the model without MCL and SVR modules, 3.50 % compared to the model without SVR, and 2.83 % compared to the model without MCL. The results also demonstrate that the SVR or MCL module can improve both clinical metrics and disease classification metrics, highlighting their effectiveness.

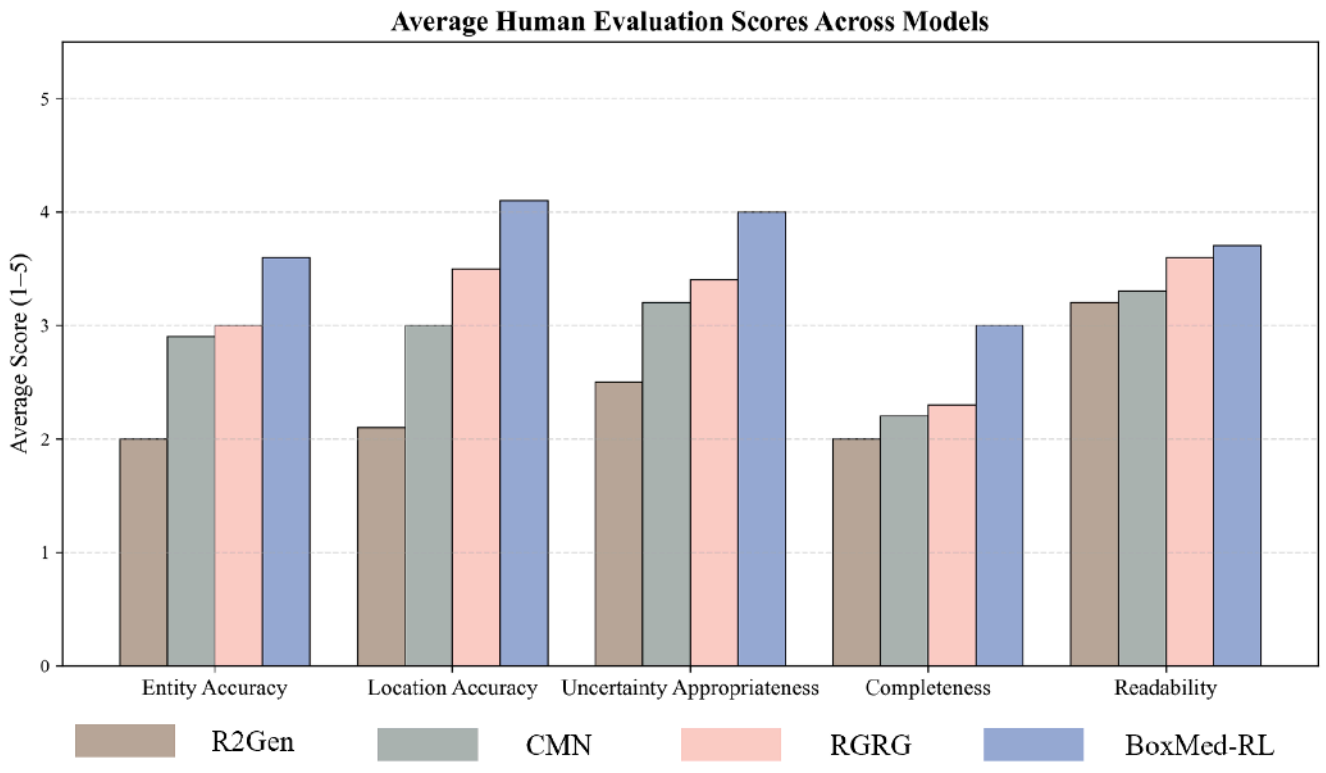

**Fig. 8.** Average 5-point Likert scores from radiologist evaluation of generated reports for reproduced models and our BoxMed-RL, across five metrics: Entity Accuracy, Location Accuracy, Uncertainty Appropriateness, Completeness, and Readability.

**Table 6**
Evaluation of reproduced methods across various evaluation metrics in CheXpert Plus. The best results are highlighted in **bold**. A one-tailed Wilcoxon signed-rank test is performed on each strategy, with $^{*}$ indicating a p-value less than 0.05 and $^{**}$ indicating a p-value less than 0.01.

| Method | BLEU-4 | METEOR | ROUGE-L | CheXbert |
| --- | --- | --- | --- | --- |
| R2Gen | 0.075$^{**}$ | 0.118$^{**}$ | 0.240 $^{**}$ | 0.195$^{**}$ |
| CMN | 0.073 $^{**}$ | 0.130 $^{**}$ | 0.249$^{**}$ | 0.236$^{**}$ |
| RGRG | 0.069 $^{**}$ | 0.124$^{**}$ | 0.237$^{**}$ | 0.261$^{**}$ |
| **Our BoxMed-RL** | **0.094** | **0.147** | **0.264** | **0.311** |

### 4.4.3. Impacts of different training strategies

Within the BoxMed-RL framework, it primarily consists of a vision tower, a component of the large vision-language model where the image is encoded; a projector, which is used to project the encoded image features for the subsequent language model to process; and a language model to decode these inputs for meaningful text generation. In this section, we evaluated various training strategies for BoxMed-RL, including: (1) freezing either the vision tower or the projector, and (2) incorporating an adapter into the language model. From Table 5, the training strategies for adapting report generation tasks produce comparable CheXbert scores, demonstrating that the generated reports closely align with the clinical semantic content of ground truth reports. This consistency highlights the robustness of the Pretraining Phase. However, with the inclusion of an adapter, BoxMed-RL generates better-formatted reports, achieving higher scores on NLG metrics. Notably, the best results were achieved without training the vision tower or projector. This outcome likely stems from the pretraining phase, where the vision tower and projector already developed the ability to associate sentences with visual features. Fine-tuning them may lead to performance degradation.

## 4.5. Case studies

To illustrate the strengths and limitations of BoxMed-RL, we present case studies covering both successful and failed scenarios (Fig. 6, Fig. 7). These span clinically relevant findings, from subtle common abnormalities (e.g., mild cardiomegaly, pulmonary edema) to rare or ambiguous ones (e.g., subcutaneous emphysema, posterior rib fractures). Each case is evaluated through generated reports, reasoning chains, and anatomical grounding, revealing BoxMed-RL's behavior under accurate and erroneous conditions.

In Fig. 6, we show four representative successes: mild cardiomegaly and pulmonary edema (a), presence of a chest port (b), a left-sided chest tube with normal heart size (c), and mild basilar opacities with consolidation (d). In each, BoxMed-RL correctly identifies findings, generates radiologist-like reasoning, and localizes the relevant anatomy. For instance, in the first case it links "mild cardiomegaly" to heart size and focuses on the cardiac silhouette; in the second and third, it detects support devices and localizes the mediastinum or left lung; in the fourth, it associates "retrocardiac opacity" with consolidation near the left lung base. These examples show that BoxMed-RL produces linguistically aligned, spatially grounded reports with interpretable reasoning. Such transparency fosters trust and enables understanding of how individual report statements arise from image evidence.

Conversely, Fig. 7 shows four challenging failures: posterior rib fractures (a,b), subcutaneous emphysema (c), and small pleural effusion (d). In the first two, fractures are subtle and low contrast, leading the model to attend correctly to ribs or spine but miss the lesion. In the third, BoxMed-RL inspects the pleural region but misclassifies subcutaneous emphysema as pleural pathology, likely due to its rarity and visual overlap. In the fourth, a faint effusion is missed, reflecting its low contrast and limited representation in training. Even in failure, BoxMed-RL exhibits clinically coherent reasoning and aligned region grounding rather than arbitrary predictions. This behavior supports verification and error analysis, reinforcing its potential for trustworthy clinical decision support.

Beyond individual examples, we further quantify these patterns through a stratified error analysis, which summarizes BoxMed-RL's F1-scores relative to their label frequencies in the MIMIC test set (Table S2, supplementary material). Consistent with the qualitative cases, performance degrades notably for rare or small-region findings (e.g., Fracture, Pleural Other, Lung Lesion). A Spearman correlation between F1-score and the logarithm of label frequency revealed a strong positive relationship ($\rho = 0.67$, $p < .01$), confirming that BoxMed-RL's errors concentrate in low-frequency, spatially subtle categories where data scarcity limits its strength.

## 4.6. Human evaluation and external validation

To substantiate clinical utility beyond text overlap metrics, a blinded radiologist evaluation was conducted to assess the clinical reliability of generated reports across five dimensions: Entity Accuracy, Location Accuracy, Uncertainty Appropriateness, Completeness, and Readability. Each criterion was rated on a 5-point Likert scale from Poor to Excellent. As shown in Fig. 8, BoxMed-RL achieved the highest average ratings across all metrics, particularly in Entity Accuracy and Location Accuracy, reflecting enhanced medical term coverage and spatial awareness attributed to the proposed MCL and SVR modules. The detailed evaluation details and a representative example are provided in the supplementary material (Fig. S4).

To evaluate robustness to scanner and patient mix shifts, an external validation is conducted on CheXpert Plus. As shown in Table 6, BoxMed-RL consistently outperforms reproduced baselines, achieving the highest scores across all evaluation metrics. In particular, BoxMed-RL improves CheXbert by an average of 36 % compared with prior models, demonstrating strong generalization and robustness to cross-domain distribution shifts.

## 4.7. Ethical considerations

While BoxMed-RL advances explainable and verifiable X-ray report generation, several ethical considerations and risks must be acknowledged. First, over-reliance on automated reports could lead clinicians to accept AI outputs uncritically, potentially reducing vigilance or even contributing to clinical deskilling over time. Second, like other generative models, BoxMed-RL may still produce hallucinations or misdiagnoses, particularly when faced with cases outside its training distribution. Such errors could have serious consequences if not carefully monitored.

To mitigate these errors, several strategies are proposed. BoxMed-RL is designed as an assistive tool rather than a replacement, with its outputs intended to support radiologists' workflow. The integration of MCL and SVR explicitly encourages the model to justify its predictions with anatomical evidence, helping clinicians verify findings rather than relying on text alone. In practice, deployment should ensure human-in-the-loop validation, where radiologists remain the final decision-makers. Additional safeguards include calibrating outputs with uncertainty estimates and continuously updating the system with post-deployment monitoring. Future extensions could incorporate calibrated uncertainty estimation to flag ambiguous or out-of-distribution cases for closer human review. Interactive interfaces that link textual statements to corresponding image regions may further enhance transparency and user trust. Maintaining post-deployment monitoring and dataset updates will also be important to ensure fairness and reliability across institutions.

### 4.8. Patient context in report generation

Although our work focuses on imaging-driven generation, where BoxMed-RL enhances transparency and verifiability, we note that radiologists in practice do not rely on images alone. The patient context, including prior imaging studies and reports, electronic health records (EHR), and radiology information systems (RIS), is important for interpretation. For example, detecting a subtle opacity may carry different implications depending on whether it represents a new lesion or a stable abnormality documented in prior exams.

The absence of patient context in BoxMed-RL means that each image is interpreted in isolation. As a result, the model cannot capture longitudinal patterns or case-specific nuances, such as distinguishing between first-time and follow-up studies or tailoring descriptions to known patient history. While this does not affect factual grounding, it may reduce the clinical plausibility of generated reports in context-sensitive scenarios. Addressing this gap would better reflect radiologists' diagnostic workflow and further strengthen the model's ability to provide case-specific interpretations.

## 5. Conclusion

In this work, we present BoxMed-RL, a novel framework for chest X-ray report generation that unifies medical concept learning with spatially verifiable reinforcement. By decomposing reports into hierarchical reasoning steps and enforcing textual–visual alignment, BoxMed-RL addresses two critical challenges: the mismatch between model outputs and radiologists' structured diagnostic workflows, and the misalignment between descriptions and visual evidence. Experiments on two public benchmarks demonstrate that BoxMed-RL consistently outperforms state-of-the-art methods across language quality, clinical accuracy, and composite evaluation metrics.

Despite these efforts, limitations remain. Our study focuses on chest X-ray datasets, which may not capture the full diversity of clinical imaging tasks. Reliance on bounding-box annotations also limits scalability, as such labels are costly and scarce for rare diseases or other modalities. We observed weaker performance on categories such as Lung Lesion, Pleural Other, and Fracture, which could be mitigated by curating more diverse, fine-grained grounding annotations. Although the MCL dataset was curated under senior radiologist supervision, formal multi-rater validation and inter-rater reliability analysis remain future steps to strengthen the credibility and consistency of the reasoning chains. Future work will extend BoxMed-RL to CT and MRI, explore multi-modal and 3D settings, and reduce dependence on dense annotations via semi- or weakly supervised learning. Validation in real-world clinical workflows will also be essential to assess robustness and safety. Another key direction is integrating patient context to better emulate radiologists' decision-making and enhance report quality.

## CRediT authorship contribution statement

**Peiyuan Jing:** Writing – review & editing, Writing – original draft, Visualization, Validation, Supervision, Software, Resources, Methodology, Formal analysis, Data curation, Conceptualization; **Kinhei Lee:** Writing – review & editing, Writing – original draft, Visualization, Validation, Methodology, Formal analysis, Data curation, Conceptualization; **Zhenxuan Zhang:** Writing – review & editing, Writing – original draft, Visualization, Validation, Supervision, Methodology, Formal analysis, Data curation, Conceptualization; **Huichi Zhou:** Visualization, Validation, Methodology, Conceptualization; **Zhengqing Yuan:** Software, Resources, Data curation; **Zhifan Gao:** Supervision; **Lei Zhu:** Supervision; **Giorgos Papanastasiou:** Supervision; **Yingying Fang:** Supervision; **Guang Yang:** Supervision, Resources, Project administration, Funding acquisition.

## Declaration of competing interest

The authors declare that they have no known competing financial interests or personal relationships that could have appeared to influence the work reported in this paper.

## Acknowledgment

Guang Yang was supported in part by the ERC IMI (101005122), the H2020 (952172), the MRC (MC/PC/21013), the Royal Society (IEC/NSFC/211235), the NVIDIA Academic Hardware Grant Program, the SABER project supported by Boehringer Ingelheim Ltd, NIHR Imperial Biomedical Research Centre (RDA01), The Wellcome Leap Dynamic resilience program (co-funded by Temasek Trust)., UKRI guarantee funding for Horizon Europe MSCA Postdoctoral Fellowships (EP/Z002206/1), UKRI MRC Research Grant, TFS Research Grants (MR/U506710/1), Swiss National Science Foundation (Grant No. 220785), and the UKRI Future Leaders Fellowship (MR/V023799/1, UKRI2738). Peiyuan Jing is supported by the Swiss National Science Foundation (SNSF) under grant number 20HW-1 220785. Zhenxuan Zhang is supported by a CSC Scholarship.

## Supplementary material

Supplementary material associated with this article can be found, in the online version, at 10.1016/j.media.2025.103910